\documentclass{article}

\usepackage{arxiv}
\usepackage[utf8]{inputenc} % allow utf-8 input
\usepackage[T1]{fontenc}    % use 8-bit T1 fonts
\usepackage{amsmath}
\usepackage{xcolor}
\usepackage{colortbl}       % for \rowcolor in tables
\usepackage{enumitem}       % A., B., ... labels of the method lists
\usepackage{placeins}       % \FloatBarrier
\usepackage{hyperref}       % hyperlinks
\usepackage{url}            % simple URL typesetting
\usepackage{booktabs}       % professional-quality tables
\usepackage{amsfonts}       % blackboard math symbols
\usepackage{nicefrac}       % compact symbols for 1/2, etc.
\usepackage{microtype}      % microtypography
\usepackage{graphicx}
\usepackage{doi}

\title{\upshape\textit{Bg-Mix}: On Using Image Backgrounds for Data Augmentation in Electron Microscopy Images}

\author{%
	\begin{tabular}{cccc}
		Ankur Srivastava$^{1,2,}$\textsuperscript{*} & Joana Sarah Grah$^{1}$ & Katharina Ladewig$^{1,2}$ & Nils K\"{o}rber$^{1,}$\textsuperscript{*} \\
	\end{tabular} \\[4pt]
	$^{1}$ Centre for Artificial Intelligence in Public Health Research, Robert Koch Institute, Berlin, Germany \\
	$^{2}$ Medical University Lausitz – Carl Thiem, Cottbus, Germany
}

\date{}

\renewcommand{\headeright}{}
\renewcommand{\undertitle}{}
\renewcommand{\shorttitle}{\textit{Bg-Mix}}

\hypersetup{
pdftitle={Bg-Mix: On Using Image Backgrounds for Data Augmentation in Electron Microscopy Images},
pdfauthor={Ankur Srivastava, Joana Sarah Grah, Katharina Ladewig, Nils K\"{o}rber},
pdfkeywords={Data Augmentation, Object Detection, Electron Microscopy},
}

\begin{document}
\maketitle

% footnote at the bottom of page 1, marked with *
\begingroup
\renewcommand{\thefootnote}{*}
\footnotetext{Corresponding authors: \texttt{\{srivastavaa, koerbern\}@rki.de}}
\endgroup

% IEEE notice, required while the paper is under review at IEEE (no mark)
\begingroup
\renewcommand{\thefootnote}{}
\addtolength{\footnotesep}{10pt}
\footnotetext{\hspace*{-1.8em}This work has been submitted to the IEEE for possible publication. Copyright may be transferred without notice, after which this version may no longer be accessible.}
\endgroup

\begin{abstract}

Electron microscopy (EM) serves as a cornerstone in virology enabling the visualization of viral particles and is widely used in research, diagnostics and morphological characterization of pathogens. Computer-assisted image analysis can help facilitate and speed up this labor-intensive method, however, typical EM virus datasets are small and sparsely labeled, resulting in a lack of readily available deep learning models due to training limitations. To overcome data shortages and leverage content-rich, diverse sample backgrounds, we employed background-based data augmentation. To this end, we propose \textit{Bg-Mix}, a form of data augmentation method, which decomposes training images into foreground objects and backgrounds and subsequently rearranges both into novel augmented images. This not only enhances the performance of object detection models but also increases the diversity of the training data. Using \textit{Bg-Mix} provides an increase between \textbf{+0.8} and \textbf{+3.3} box mean average precision (mAP) points over the baseline when tested across two datasets using two architectures, thereby demonstrating its effectiveness.
\end{abstract}


\section{Introduction}
\label{sec:intro}

Negative-staining Scanning-Transmission/Transmission Electron Microscopy (STEM/TEM) of clinical samples can help diagnose unknown and emerging infections \cite{goldsmith2009modern}. Recognizing virus particles in those images, however, requires expert knowledge: viruses must be differentiated from debris and cell material, and morphologically similar viruses must be distinguished from one another. Automated virus detection can reduce the workload of experts by alleviating the need for labor-intensive manual analysis, while potentially enabling faster and more accurate diagnoses.

In light of the rise of deep learning methods in imaging and microscopy data availability, a myriad of models for object detection/instance segmentation \cite{he2017mask, lyu2022rtmdet} and classification \cite{wang2017residual, woo2023convnext} have been developed over the past decade. Still, biomedical imaging datasets remain costly to acquire and particularly time-consuming to annotate, often resulting in limited dataset availability and size. Several strategies have been proposed to address these limitations, including transfer learning \cite{kora2022transfer}, synthetic image generation \cite{koetzier2024generating}, and data augmentation \cite{goceri2023medical}. In this work, we focus on the latter strategy, which is particularly well suited to negative-staining STEM/TEM images because of their content-rich backgrounds.

In recent years, several methods have been proposed for data augmentation: In the Copy-Paste method \cite{ghiasi2021simple}, objects are cut from a random training image and pasted at random on another training image. The model in \cite{zhang2023background} learns the kind of background each class appears on and pastes objects only where the background fits. In \cite{li2024simple} the labeled objects are kept intact and only the background is synthetically generated by using diffusion models \cite{lu2025dpm}.

All of the above works use large numbers of natural training images, for example from the COCO dataset \cite{lin2014microsoft}. However, STEM/TEM datasets are often limited in size, typically comprising only a few hundred annotations, making data augmentation particularly valuable for supervised learning approaches.

Here, we propose \textit{Bg-Mix}, a data augmentation method where segmented objects are extracted from one training image, an object-free background is extracted from another one and both are subsequently rearranged into a new augmented training image. We thus make use of the rich and diverse negative-stained STEM/TEM backgrounds without the need to either learn them or generate them artificially. We tested three ways to obtain object-free backgrounds, two ways to blend the object's edges and six ways to place the foreground objects. We evaluate on two public negative-stained STEM/TEM virus datasets with two model architectures including an ablation study on different \textit{Bg-Mix} modes and compare with the previously described Copy-Paste method.

\section{Method}
\label{sec:method}

\subsection{Overview}
\label{subsec:overview}

\begin{figure}[t]
  \centering
  \includegraphics[width=0.9\textwidth]{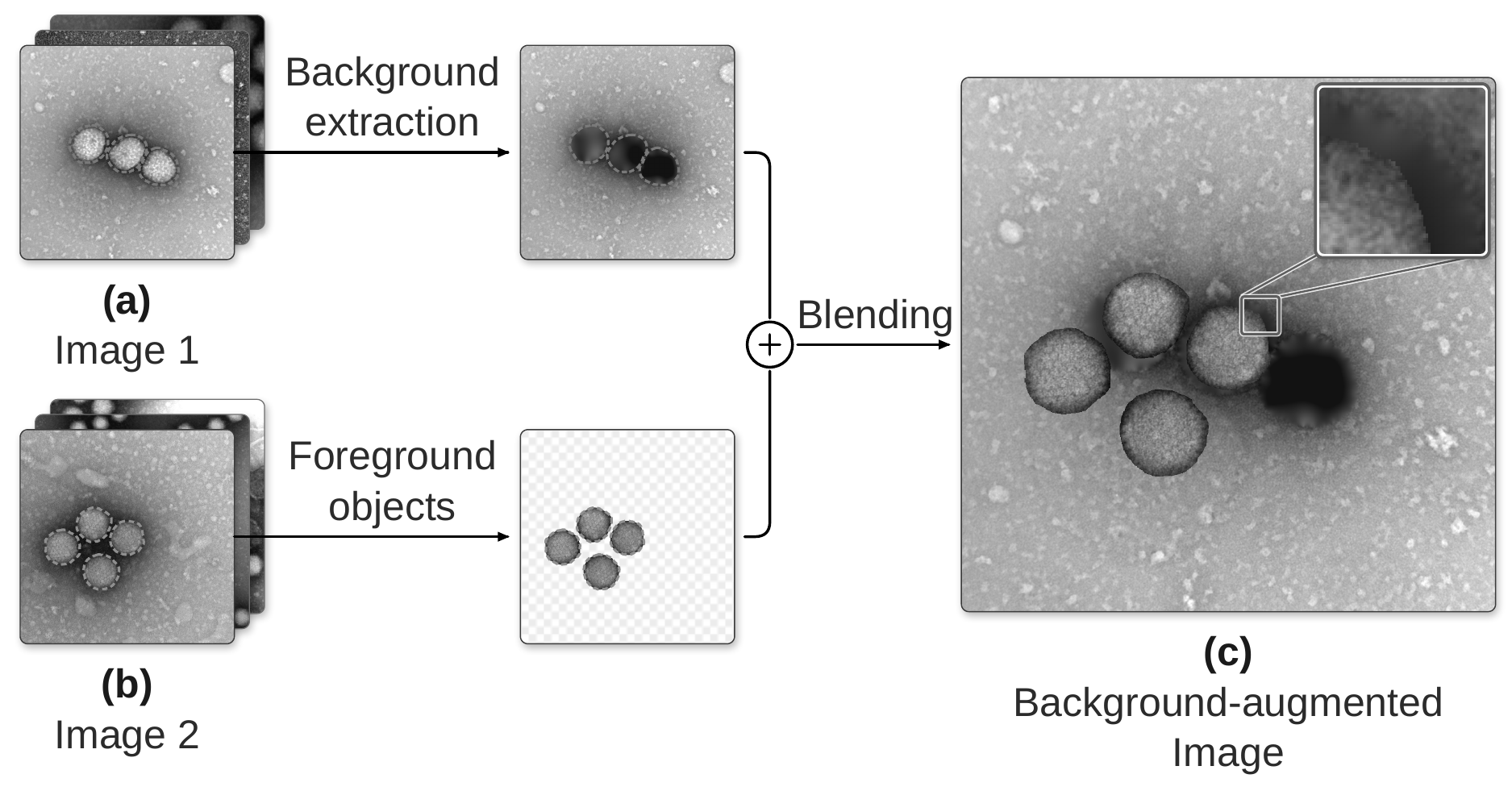}
  \caption{\textit{Bg-Mix}. The background of a random training image (a) and the foreground objects of another random training image (b) are extracted and blended to form the background-augmented image (c).}
  \label{fig:method}
\end{figure}

\textit{Bg-Mix} randomly extracts foreground objects and object-free backgrounds from training images in a given dataset and subsequently mixes them back together to form new training samples (cf. Fig. \ref{fig:method}).

\subsection{Background Extraction}
\label{subsec:backext}

From a randomly drawn training image, the background is extracted in one of the following three ways:
\begin{enumerate}[label=\Alph*.]
    \item Selecting the largest object-free square in the image and resizing it back to the image size,
    \item Removing its objects using segmentation masks and setting the pixels of the mask to zero,
    \item Removing its objects using segmentation masks and applying biharmonic inpainting \cite{damelin2018surface}, which fills the mask by smooth interpolation based on the surrounding pixel values.
\end{enumerate}

\subsection{Object Blending}
\label{subsec:obj_bl}

\textit{Bg-Mix} incorporates two modes to blend extracted foreground objects and backgrounds:
\begin{enumerate}[label=\Alph*., start=4]
    \item Hard edges: The foreground objects are pasted onto the background as they are,
    \item Blurred edges: The pasted foreground objects' boundaries are blurred using a Gaussian kernel of size $3 \times 3$ with standard deviation $\sigma = 0.8$.
\end{enumerate}

\subsection{Object Placement}
\label{subsec:obj_pl}

We consider six object placement methods depending on the foreground, background and the placement order:
\begin{enumerate}[label=\Alph*., start=6]
    \item Foreground objects from one image, hence one class, are pasted onto a background from any class at their original positions,
    \item Foreground objects from one image, hence one class, are pasted onto a background from the same class at their original positions,
    \item Foreground objects from multiple images of a single class are pasted onto a background from any class at random non-overlapping positions,
    \item Foreground objects from multiple images of a single class are pasted onto a background from the same class at random non-overlapping positions,
    \item Foreground objects from multiple images of any class are pasted onto a background from any class at random non-overlapping positions,
    \item At least one foreground object from each class is pasted onto a background from any class at random non-overlapping positions.
\end{enumerate}

\section{Experiments}
\label{sec:exp}

\providecommand{\gain}{}\renewcommand{\gain}[1]{+#1}           % above the baseline: +
\providecommand{\loss}{}\renewcommand{\loss}[1]{\textminus#1}  % below the baseline: minus
\definecolor{Bg-Mixrow}{gray}{0.93}  % slight grey band for the rows (needs \usepackage{colortbl})

\begin{table}[b]
  \centering
  \caption{Mask R-CNN and RTMDet-Ins models with Copy-Paste and \textit{Bg-Mix} on EMViR and VISN datasets. The brackets show the change from the first row of the same block. $p$ is the probability with which Copy-Paste or \textit{Bg-Mix} was applied to a training image ($p{=}0.5$ for EMViR; $p{=}0.01$ for VISN). Best results are shown in bold.}
  \label{tab:Bg-Mix_emvir_visn}
  \vspace{5pt}
  \renewcommand{\arraystretch}{1.3}
\begin{tabular}{l|c|c}
    %\hline
    \multicolumn{1}{c|}{\raisebox{-0.5\normalbaselineskip}[0pt][0pt]{Model}} & EMViR & VISN \\
                       & box mAP & box mAP \\
    \hline
    Mask R-CNN & 55.7 & 77.7 \\
    \quad + Copy-Paste ($p{=}0.5$ / $0.01$) & 55.5 (\textminus0.2) & 75.9 (\textminus1.8) \\
    \rowcolor{Bg-Mixrow}
    \quad \textbf{+ \textit{Bg-Mix}} ($p{=}0.5$ / $0.01$) & \textbf{58.3 (\gain{2.6})} & \textbf{78.5 (\gain{0.8})} \\
    \hline
    RTMDet-Ins & 54.5 & 57.9 \\
    \quad + Copy-Paste ($p{=}0.5$ / $0.01$) & 56.7 (+2.2) & \textbf{60.5 (+2.6)} \\
    \rowcolor{Bg-Mixrow}
    \quad \textbf{+ \textit{Bg-Mix}} ($p{=}0.5$ / $0.01$) & \textbf{57.8 (\gain{3.3})} & 58.7 (\gain{0.8}) \\
    %\hline
  \end{tabular}
\end{table}

\providecommand{\gain}{}\renewcommand{\gain}[1]{+#1}           % above the baseline: +
\providecommand{\loss}{}\renewcommand{\loss}[1]{\textminus#1}  % below the baseline: minus
\providecommand{\cm}{\ifdefined\checkmark\checkmark\else\ensuremath{\surd}\fi}

\begin{table}[t]
  \centering
  \caption{Ablation study of \textit{Bg-Mix} on the EMViR dataset with Mask R-CNN: all combinations of background extraction, blending and object placement methods. All runs except baseline use $p = 0.5$. Best setup is shown in bold.}
  \label{tab:Bg-Mix_all36_ticks_emvir_maskrcnn}
  \renewcommand{\arraystretch}{1.05}
  \vspace{5pt}
  \begin{tabular}{c|ccc|cc|cccccc|c}
      & \multicolumn{3}{c|}{Extraction}
      & \multicolumn{2}{c|}{Blending} & \multicolumn{6}{c|}{Placement}
      & \\
    Run & \makebox[8pt]{A} & \makebox[8pt]{B} & \makebox[8pt]{C} & \makebox[12pt]{D} & \makebox[12pt]{E} & F & G & H & I & J & K & box mAP \\
    \hline
    Baseline
      & \multicolumn{3}{c|}{--} & \multicolumn{2}{c|}{--} & \multicolumn{6}{c|}{--} & 55.7\phantom{ (\gain{2.0})} \\
    \hline
     1 & \cm &     &     & \cm &     & \cm &     &     &     &     &     & 57.7 (\gain{2.0}) \\
     2 & \cm &     &     &     & \cm & \cm &     &     &     &     &     & 57.0 (\gain{1.3}) \\
     3 &     & \cm &     & \cm &     & \cm &     &     &     &     &     & 57.9 (\gain{2.2}) \\
     4 &     & \cm &     &     & \cm & \cm &     &     &     &     &     & 57.5 (\gain{1.8}) \\
     5 &     &     & \cm & \cm &     & \cm &     &     &     &     &     & \textbf{58.3 (\gain{2.6})} \\
     6 &     &     & \cm &     & \cm & \cm &     &     &     &     &     & 57.6 (\gain{1.9}) \\
    \hline
     7 & \cm &     &     & \cm &     &     & \cm &     &     &     &     & 56.1 (\gain{0.4}) \\
     8 & \cm &     &     &     & \cm &     & \cm &     &     &     &     & 56.7 (\gain{1.0}) \\
     9 &     & \cm &     & \cm &     &     & \cm &     &     &     &     & 56.7 (\gain{1.0}) \\
    10 &     & \cm &     &     & \cm &     & \cm &     &     &     &     & 57.1 (\gain{1.4}) \\
    11 &     &     & \cm & \cm &     &     & \cm &     &     &     &     & 56.7 (\gain{1.0}) \\
    12 &     &     & \cm &     & \cm &     & \cm &     &     &     &     & 58.1 (\gain{2.4}) \\
    \hline
    13 & \cm &     &     & \cm &     &     &     & \cm &     &     &     & 55.9 (\gain{0.2}) \\
    14 & \cm &     &     &     & \cm &     &     & \cm &     &     &     & 53.8 (\loss{1.9}) \\
    15 &     & \cm &     & \cm &     &     &     & \cm &     &     &     & 58.0 (\gain{2.3}) \\
    16 &     & \cm &     &     & \cm &     &     & \cm &     &     &     & 55.7 ($\pm$0.0) \\
    17 &     &     & \cm & \cm &     &     &     & \cm &     &     &     & 56.5 (\gain{0.8}) \\
    18 &     &     & \cm &     & \cm &     &     & \cm &     &     &     & 56.4 (\gain{0.7}) \\
    \hline
    19 & \cm &     &     & \cm &     &     &     &     & \cm &     &     & 55.7 ($\pm$0.0) \\
    20 & \cm &     &     &     & \cm &     &     &     & \cm &     &     & 55.7 ($\pm$0.0) \\
    21 &     & \cm &     & \cm &     &     &     &     & \cm &     &     & 57.0 (\gain{1.3}) \\
    22 &     & \cm &     &     & \cm &     &     &     & \cm &     &     & 56.6 (\gain{0.9}) \\
    23 &     &     & \cm & \cm &     &     &     &     & \cm &     &     & 55.7 ($\pm$0.0) \\
    24 &     &     & \cm &     & \cm &     &     &     & \cm &     &     & 57.2 (\gain{1.5}) \\
    \hline
    25 & \cm &     &     & \cm &     &     &     &     &     & \cm &     & 56.6 (\gain{0.9}) \\
    26 & \cm &     &     &     & \cm &     &     &     &     & \cm &     & 56.0 (\gain{0.3}) \\
    27 &     & \cm &     & \cm &     &     &     &     &     & \cm &     & 54.3 (\loss{1.4}) \\
    28 &     & \cm &     &     & \cm &     &     &     &     & \cm &     & 56.0 (\gain{0.3}) \\
    29 &     &     & \cm & \cm &     &     &     &     &     & \cm &     & 58.1 (\gain{2.4}) \\
    30 &     &     & \cm &     & \cm &     &     &     &     & \cm &     & 55.4 (\loss{0.3}) \\
    \hline
    31 & \cm &     &     & \cm &     &     &     &     &     &     & \cm & 57.8 (\gain{2.1}) \\
    32 & \cm &     &     &     & \cm &     &     &     &     &     & \cm & 55.9 (\gain{0.2}) \\
    33 &     & \cm &     & \cm &     &     &     &     &     &     & \cm & 57.8 (\gain{2.1}) \\
    34 &     & \cm &     &     & \cm &     &     &     &     &     & \cm & 57.5 (\gain{1.8}) \\
    35 &     &     & \cm & \cm &     &     &     &     &     &     & \cm & 56.3 (\gain{0.6}) \\
    36 &     &     & \cm &     & \cm &     &     &     &     &     & \cm & 56.9 (\gain{1.2}) \\
  \end{tabular}
\end{table}

\subsection{Datasets}
\label{subsec:datasets}

For virus segmentation, we utilized two publicly available datasets in this study. In both datasets, each image contains objects of only one class, which enabled us to use class-matched backgrounds for experiments.

The EMViR dataset \cite{klimczok2025emvir} contains 1084 negative-staining STEM images with segmentation masks of four classes of respiratory viruses (Adeno, Corona, Orthomyxo and Paramyxo), split into 762 training, 157 validation and 165 test images.

The virus dataset of \cite{xiao2023visn}, abbreviated VISN in the following, contains 622 negative-staining TEM images with segmentation masks of six virus classes (SARS-CoV-2, H1N1, Respiratory syncytial, Herpes simplex, Adeno and Vaccinia), divided into 499 training and 123 test images.

As pre-processing steps, (1) the EMViR object masks were converted to COCO json annotations to match the VISN format, (2) a VISN validation set was created from a stratified 15\% of the original training set, and (3) all images across both datasets were resized to $1024 \times 1024$ pixels and their annotations were rescaled proportionally.

\subsection{Training}
\label{training}

We evaluated \textit{Bg-Mix} on both datasets with two object detection/instance segmentation architectures: Mask R-CNN \cite{he2017mask} with a ResNet-50-FPN backbone and RTMDet-Ins-L \cite{lyu2022rtmdet}. Both models were initialized with weights pretrained on COCO \cite{lin2014microsoft}, implemented using the MMDetection toolbox \cite{chen2019mmdetection} and fine-tuned on each dataset. We kept each architecture's published optimizer and learning rate schedule and extended it to 500 epochs: SGD with momentum 0.9 and weight decay $10^{-4}$, with the learning rate divided by 10 at epochs 333 and 458, for Mask R-CNN, and AdamW with weight decay 0.05 and cosine decay from epoch 250 for RTMDet-Ins. With a uniform batch size of 8, we scaled the published learning rates linearly: from 0.02 (batch size 16) to 0.01 for Mask R-CNN, and from 0.004 (batch size 256) to $1.25 \times 10^{-4}$ for RTMDet-Ins.

\textit{Bg-Mix} was applied independently to each training image with probability $p$, so the number of training images per epoch stayed fixed. To observe the effect of \textit{Bg-Mix} alone, no other augmentation was used. We compared \textit{Bg-Mix} using the same architectures without augmentation (baseline) and with Copy-Paste \cite{ghiasi2021simple}, the latter applied with the same probability as \textit{Bg-Mix}: $p = 0.5$ on EMViR and $p = 0.01$ on VISN. This choice was made based on our observation that for the EMViR dataset, $0.25 \le p \le 0.75$ increased performance, but for the VISN dataset, $p>0.01$ decreased the performance. We evaluated on the validation set every 10 epochs and kept the checkpoint with the best validation box mAP for testing. All runs were averaged over three random seeds.

\subsection{Results}
\label{results}

\textit{Bg-Mix} obtained the largest gain on the EMViR dataset for both architectures (+2.6 and +3.3 box mAP) (cf. Table \ref{tab:Bg-Mix_emvir_visn}). For the VISN dataset, \textit{Bg-Mix} had a small increase (+0.8 and +0.8 box mAP) for both models. The small increase for \textit{Bg-Mix} (and the corresponding decrease for Copy-Paste) for the Mask R-CNN architecture and the VISN dataset can be attributed to the already saturated baseline (77.7 box mAP).

From the ablation study involving the Mask R-CNN architecture and the EMViR dataset (cf. Table \ref{tab:Bg-Mix_all36_ticks_emvir_maskrcnn}), we observe the following, reporting gains of box mAP in points averaged over all runs using the corresponding option: (1) Background Extraction: inpainting ($+1.2$) and zeroing the object pixels ($+1.1$) perform best, (2) Blending: hard edges are slightly better than blurred edges ($+1.1$ vs.\ $+0.8$), and (3) Object Placement: original objects at their original positions work best ($+2.0$). The best run combines inpainting, hard edges and original placement (run~5, 58.3 box mAP); the same setting is used for \textit{Bg-Mix} in cf. Table \ref{tab:Bg-Mix_emvir_visn}.

\section{Conclusion}
\label{sec:conclusion}

This study describes \textit{Bg-Mix}, a novel background augmentation method for virus detection in negative-staining STEM/TEM images: it extracts the objects from one training image, the object-free background from another and pastes the extracted objects onto this extracted background.

Compared to the baseline, the performance of models improved significantly by incorporating the \textit{Bg-Mix} method. Our study is limited to object detection of viruses in EM images, however, we believe our findings can be transferred to other modalities with target objects on content-rich backgrounds.

\section{Compliance with ethical standards}
\label{sec:ethics}

This study uses publicly available electron microscopy datasets of virus particles produced in vitro (EMViR \cite{klimczok2025emvir}, VISN \cite{xiao2023visn}). No ethical approval was required.

\section{Acknowledgments}
\label{sec:acknowledgments}

Conflicts of Interest: This work was supported by institutional funding from Robert Koch Institute. The authors received no specific third-party grants or external financial support for this research.

\FloatBarrier % place all tables before the references
\bibliographystyle{IEEEbib}
\bibliography{refs}

\end{document}